\documentclass{article}

\PassOptionsToPackage{numbers, compress}{natbib}

\usepackage[preprint]{neurips_2026}

\usepackage[utf8]{inputenc}
\usepackage[T1]{fontenc}
\usepackage{hyperref}
\usepackage{url}
\usepackage{booktabs}
\usepackage{amsfonts}
\usepackage{nicefrac}
\usepackage{microtype}
\usepackage{xcolor}

\usepackage{amsmath}
\usepackage{amssymb}
\usepackage{amsthm}
\usepackage{bm}
\usepackage{graphicx}
\usepackage{subcaption}
\usepackage{multirow}
\usepackage{enumitem}

\title{MissHyper: Restoring Clinical Synchronicity in Missingness-Guided Hypergraph Forecasting}

\author{
    Mingyi Ma \\
    Academy of Advanced Interdisciplinary Studies \\
    Wuhan University \\
    \texttt{money.yi@whu.edu.cn}
    \And
    Qingxiong Tan \\
    Institute for Mathematics \& Artificial Intelligence \\
    Wuhan University \\
    \texttt{tanqingxiong@whu.edu.cn}
}

\begin{document}

\maketitle

\begin{abstract}
Clinical irregular multivariate time series are shaped not only by physiological dynamics but also by the measurement process that determines when and what to observe. In event-centric models, however, co-timestamp structure can be flattened too early: measurements acquired at the same timestamp are embedded as isolated nodes, leaving local patient-state context unavailable until later message-passing layers. We study this pre-propagation representation bottleneck and address it by restoring co-timestamp context before message passing begins. We propose \textbf{MissHyper}, a \textbf{missingness-guided hypergraph} forecasting model with pre-propagation synchronicity restoration. MissHyper augments each event with a local support-density cue, aggregates co-timestamp records to recover patient-state context, and uses a missingness-guided gate to adaptively fuse node-specific evidence with the recovered context. Across PhysioNet 2012, MIMIC-III, and MIMIC-IV, MissHyper achieves consistent gains in multi-step forecasting and outperforms a strong hypergraph baseline. These results suggest that improving event initialization can benefit sparse clinical forecasting without requiring a redesigned downstream propagation architecture. Ablations indicate that snapshot restoration, adaptive fusion, and support-density encoding all contribute, pointing to event initialization as a critical design axis for sparse clinical forecasting.
\end{abstract}

\section{Introduction}
\label{sec:introduction}

Clinical forecasting often relies on irregular multivariate records, where patient state is reflected by sparse and partially observed measurements such as vital signs, laboratory tests, and treatment-related indicators. Forecasting future values from these records supports early warning, risk monitoring, and resource planning in intensive care units. Public benchmarks such as PhysioNet 2012, MIMIC-III, and MIMIC-IV have made this setting a standard testbed for clinical temporal modeling \citep{Goldberger2000PhysioNet,Silva2012PhysioNetChallenge,Johnson2016MIMICIII,Johnson2023MIMICIV}. Prior benchmarking studies have also shown that preprocessing choices and task definitions can strongly affect clinical time-series evaluation \citep{Purushotham2018BenchmarkingHealthcare,Harutyunyan2019MultitaskClinical,Wang2020MIMICExtract}. Unlike sensor streams collected at fixed frequencies, clinical observations are shaped by care routines, patient severity, and physician decisions. Missingness can therefore be informative rather than purely random \citep{Rubin1976InferenceMissing,Schafer2002MissingData,Little2002StatisticalMissingData}, and local observation density can serve as a cue for estimating the reliability of an event.

\begin{figure}[t]
    \centering
    \begin{subfigure}[t]{0.5\linewidth}
        \centering
        \includegraphics[width=\linewidth]{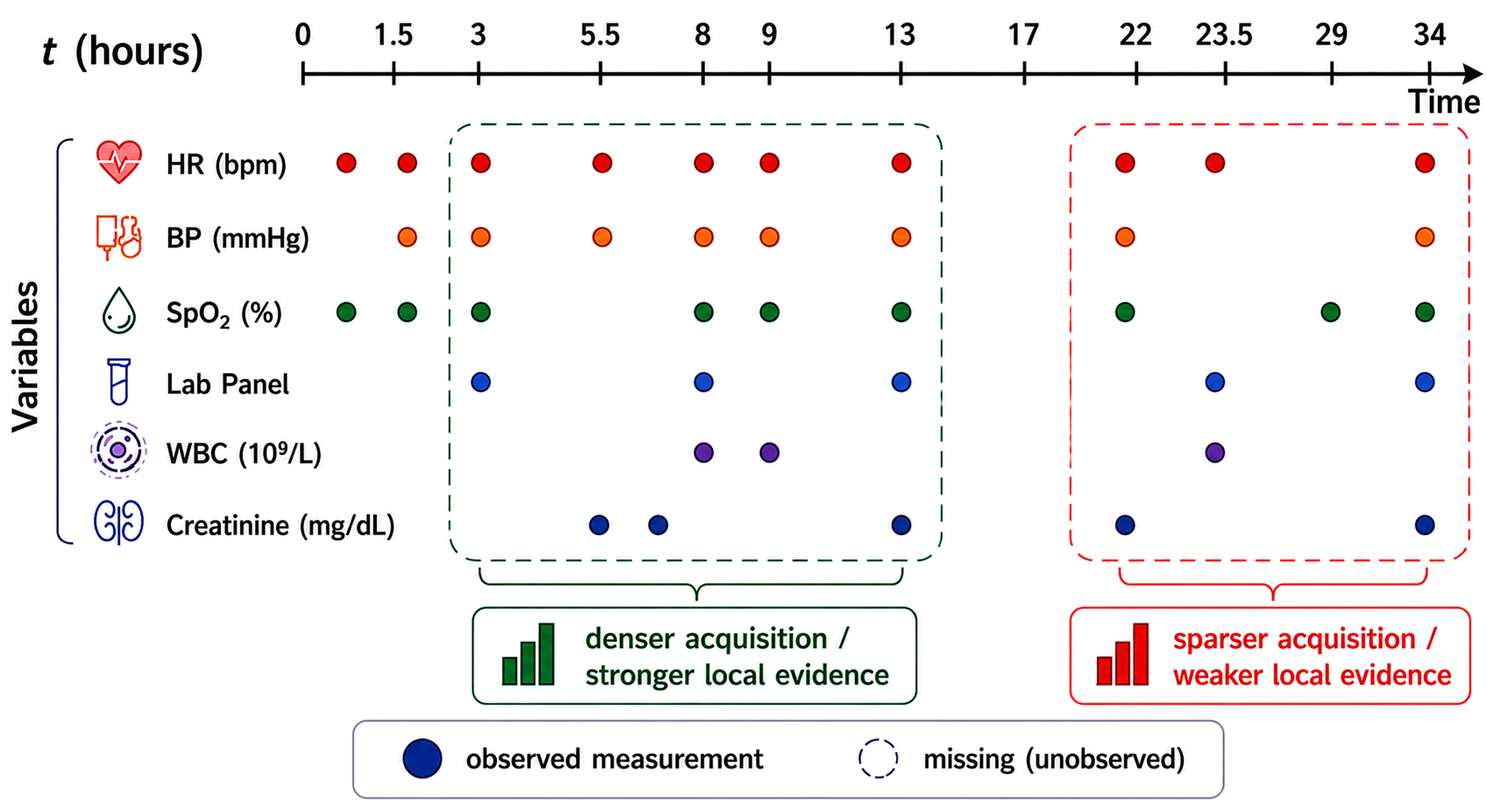}
        \label{fig:motivation_problem}
    \end{subfigure}
    \hfill
    \begin{subfigure}[t]{0.48\linewidth}
        \centering
        \includegraphics[width=\linewidth]{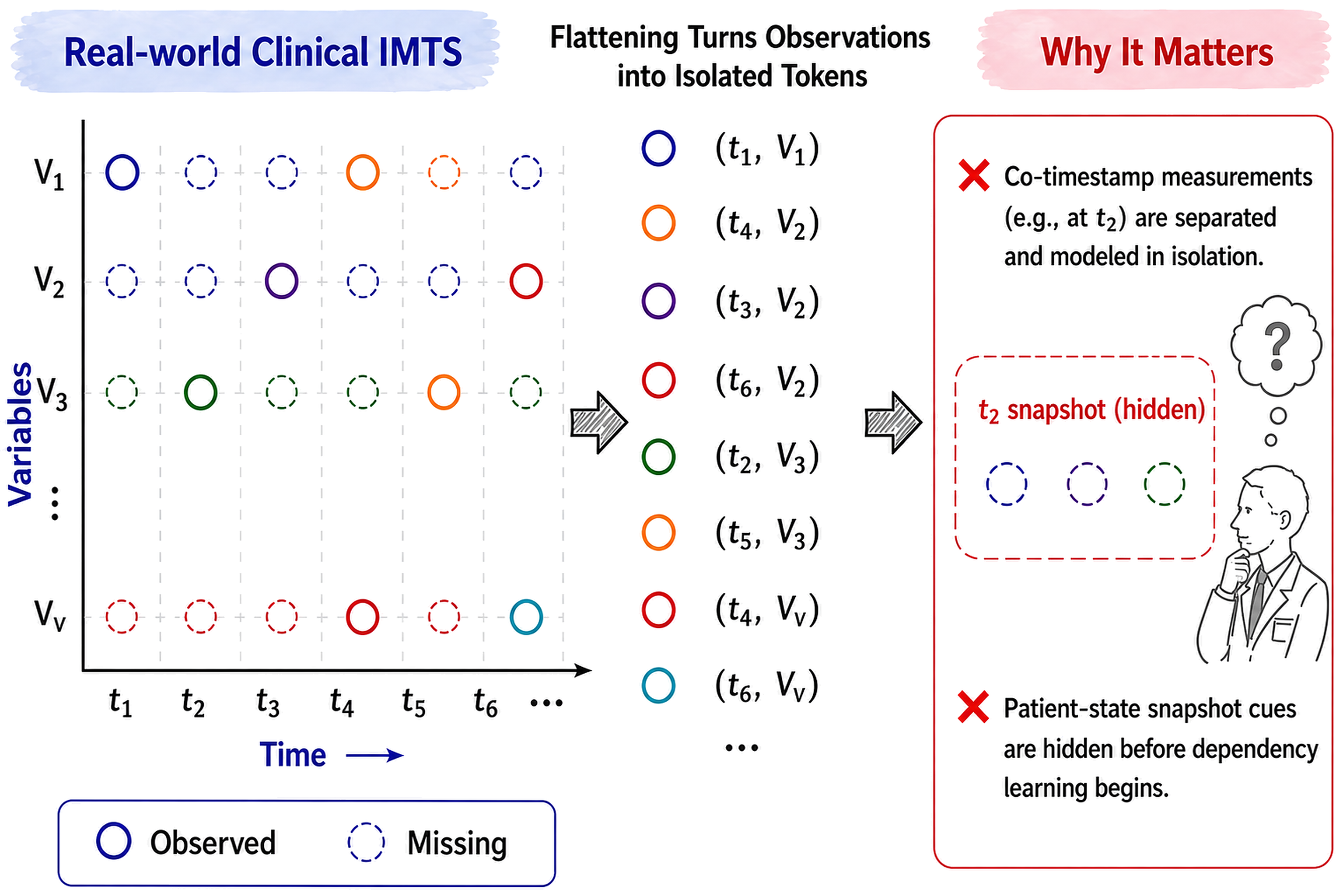}
        \label{fig:motivation_idea}
    \end{subfigure}
    \vspace{-8pt}
    \caption{Motivation and core idea of MissHyper. (a) Clinical observations are sparse, asynchronous, and unevenly supported. (b) MissHyper restores timestamp-level synchronicity before hypergraph propagation and adaptively fuses the restored context into event nodes.}
    \label{fig:motivation}
    \vspace{-35pt}
\end{figure}

This acquisition process leads to two related modeling issues in clinical IMTS. First, variables are measured at different frequencies: vital signs may be charted repeatedly, whereas laboratory tests are often ordered only when clinically needed \citep{Tipirneni2022STraTS,Zhang2023Warpformer,Huang2024DNAT,Liu2025TimeCHEAT}. Second, measurements collected at the same timestamp can form a patient-state snapshot, such as a set of concurrently charted vital signs or a laboratory panel \citep{Johnson2023MIMICIV,Wornow2023EHRSHOT,Steinberg2024MOTOR}. For instance, a moderately elevated pulse may carry different clinical implications when accompanied by low blood pressure and reduced oxygen saturation than when recorded in isolation. Event-centric IMTS models preserve sparse observations efficiently by treating records as nodes or tokens, but this design can introduce a \emph{pre-propagation representation bottleneck}: co-timestamp records that jointly describe patient state are initialized independently before they can exchange information \citep{Horn2020SeFT,Chowdhury2023PrimeNet,Yalavarthi2024GraFITi,Zhang2024TPatchGNN,Hu2025VIMTS,Luo2025HiPatch,Li2025HyperIMTS}. Figure~\ref{fig:motivation} illustrates this motivation: panel (a) shows sparse and unevenly supported clinical observations, while panel (b) shows how same-timestamp context can be restored before propagation.

Prior IMTS models have mainly improved how sparse observations are processed after their initial event representations are formed. Some methods use Gaussian processes or continuous-time dynamics \citep{Li2015MEGKernel,Futoma2017SepsisMGPRNN,Rubanova2019LatentODE,DeBrouwer2019GRUODEBayes,Chen2023ContiFormer}. Other methods rely on recurrent, attention-based, or patch-based updates \citep{Che2018GRUD,Shukla2021MTAN,Horn2020SeFT,Tipirneni2022STraTS,Huang2024DNAT,Liu2025TimeCHEAT,Hu2025VIMTS}. Recent work further explores graph or hypergraph propagation \citep{Zhang2022Raindrop,Yalavarthi2024GraFITi,Zhang2024TPatchGNN,Gravina2024TemporalGraphODE,Luo2025HiPatch,Li2025HyperIMTS}. However, even when timestamp hyperedges provide a path for later message passing, the node states entering the first propagation layer are still initialized from local event features alone. Since same-timestamp co-occurrence is directly available from the event layout, postponing it to later layers forces the model to reconstruct a simple clinical relation that could be exposed before propagation.

We propose \textbf{MissHyper}, a missingness-guided initialization strategy for hypergraph forecasting of clinical IMTS. MissHyper operates before hypergraph propagation by modifying the event representations passed into the forecasting backbone. Each event node receives a value, a role indicator, and a local support-density cue derived from the observation mask. Events sharing a timestamp are summarized into compact timestamp-level context, which is adaptively fused with node representations through a missingness-guided gate. Thus, MissHyper exposes co-timestamp clinical context before relational propagation while preserving sparse event modeling.

The main contributions of this work are summarized as follows:
\begin{itemize}
    \item We identify a pre-propagation bottleneck in sparse clinical event representations: records acquired at the same timestamp are often initialized independently even though they jointly form a patient-state snapshot. This framing shifts attention from only improving later dependency propagation to improving the information available in each event representation before propagation begins.
    \item We design MissHyper as a lightweight snapshot-aware initialization module that combines mask-derived support encoding, timestamp-level context restoration, and missingness-guided gated fusion. It restores co-timestamp context that is directly observable from the sparse event layout and uses local observation density as a reliability cue for event initialization.
    \item Experiments on PhysioNet 2012, MIMIC-III, and MIMIC-IV show consistent reductions in both MSE and MAE compared with representative irregular time-series baselines and the direct hypergraph baseline. Ablation and sensitivity analyses support the roles of snapshot restoration, adaptive fusion, and local support density, suggesting that event initialization is important when sparse clinical measurements must be interpreted together with their acquisition context.
\end{itemize}

\section{Related Work}
\label{sec:related_work}

\subsection{Sparse Clinical Sequence Modeling}
Prior IMTS studies mainly address sparse and non-uniform sampling through padding, interpolation, continuous-time modeling, recurrent updates, attention, patching, or set/event representations \citep{Shukla2021SurveyIrregularTS,Shukla2019InterpolationPrediction,KloterGens2025PhysiomeODE}. Gaussian-process methods model irregular observations with uncertainty but can be sensitive to kernel design and costly in high-dimensional clinical settings \citep{Li2015MEGKernel,Li2016GPAdapter,Futoma2017SepsisMGPRNN}. Recurrent models such as T-LSTM, Phased LSTM, and GRU-D incorporate elapsed time, masks, or decayed memory \citep{Neil2016PhasedLSTM,Baytas2017TLSTM,Che2018GRUD}. Continuous-time models describe latent dynamics between observations through ODE-based updates \citep{Chen2018NeuralODE,Rubanova2019LatentODE,DeBrouwer2019GRUODEBayes,Chen2023ContiFormer} or related flow and controlled-differential formulations \citep{Kidger2020NeuralCDE,Bilos2021NeuralFlows,Schirmer2022CRU}. Attention and event-based methods such as SeFT, mTAN, STraTS, PrimeNet, DNA-T, TimeCHEAT, and VIMTS avoid simple dense alignment and learn from sparse observed records \citep{Horn2020SeFT,Shukla2021MTAN,Tipirneni2022STraTS,Chowdhury2023PrimeNet,Huang2024DNAT,Liu2025TimeCHEAT,Hu2025VIMTS}. Imputation-oriented models use recurrence, adversarial learning, or graph propagation \citep{Yoon2017MRNN,Cao2018BRITS,Luo2019E2GAN,Cini2022GRIN}, while diffusion and self-attention variants reconstruct missing values with different generative or attention mechanisms \citep{Tashiro2021CSDI,Du2023SAITS}. These methods improve irregular temporal modeling, but they usually treat same-timestamp clinical measurements as context to be learned later or implicitly. They therefore do not explicitly address whether each sparse event should already contain local snapshot information before dependency learning begins.

\subsection{Graph and Hypergraph Dependency Learning}
Graph-based methods introduce structural bias for variable and temporal dependency learning. General graph neural networks and graph attention methods provide the basic message-passing machinery \citep{Scarselli2009GNN,Defferrard2016ChebNet,Kipf2017GCN,Velickovic2018GAT}, and later variants extend this machinery through sampling or edge-conditioned messages \citep{Hamilton2017GraphSAGE,Gilmer2017MPNN}. Time-series graph models have been used to capture inter-series or spatio-temporal dependencies \citep{Wu2020MTGNN,Yi2023FourierGNN,Han2024BigST}. For irregularly sampled series, Raindrop, Warpformer, GraFITi, Temporal Graph ODEs, and Hi-Patch learn from sparse, transformed, continuous-time, or multi-scale graph structures \citep{Zhang2022Raindrop,Zhang2023Warpformer,Yalavarthi2024GraFITi,Gravina2024TemporalGraphODE,Luo2025HiPatch}. Other recent variants combine patching, neural flows, or knowledge guidance with graph reasoning \citep{Zhang2024TPatchGNN,Mercatali2024GNeuralFlow,Luo2024KEDGN}. Hypergraph methods further allow one hyperedge to connect multiple records \citep{Feng2019HGNN,Jiang2019DHGNN,Yadati2019HyperGCN}. Hypergraph attention and structure-learning variants enrich this representation \citep{Zhang2020HyperSAGNN,Bai2021HypergraphConvolutionAttention,Cai2022HSL,Gao2023HGNNPlus}. 
HyperIMTS applies timestamp and variable hyperedges to IMTS forecasting \citep{Li2025HyperIMTS}, while recent hypergraph variants further explore adaptive or multi-scale structures for time-series forecasting \citep{Shang2024AdaMSHyper}. These approaches strengthen dependency propagation and avoid excessive padding, but their initial node states can still be formed mainly from local event features. MissHyper is motivated by this limitation: before graph or hypergraph propagation, it restores directly observed co-timestamp context and uses missingness-derived support density to decide how strongly that context should shape each event representation.

\section{Proposed Method}
\label{sec:method}

\begin{figure*}[t]
    \centering
    \includegraphics[width=1\textwidth]{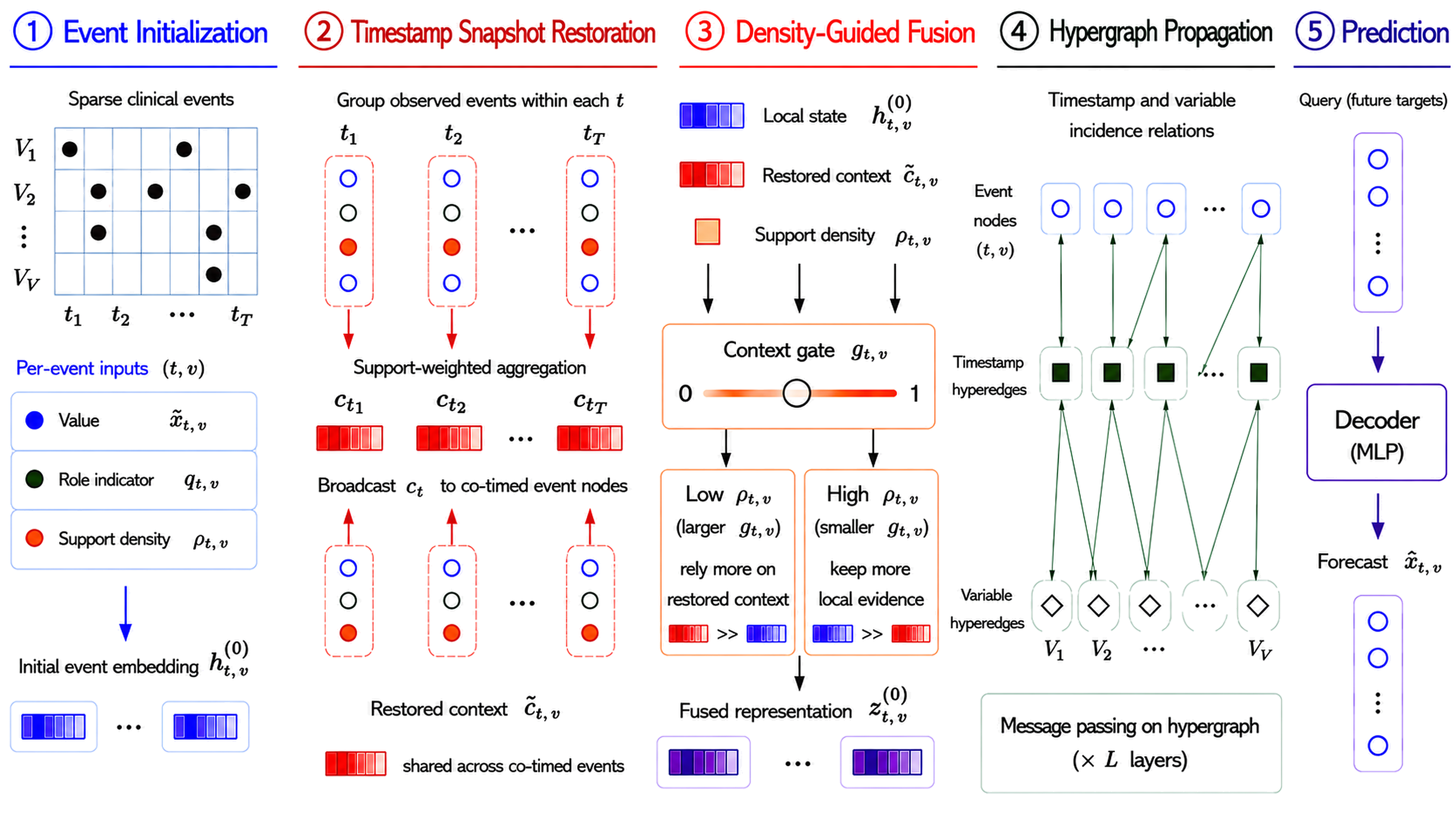}
    \vspace{-15pt}
    \caption{Overall framework of MissHyper: restoring co-timestamp clinical snapshots before hypergraph propagation. MissHyper first encodes mask-derived local support density for each sparse event, then aggregates events observed at the same timestamp to recover patient-state context, and finally uses a missingness-guided gate to fuse node-specific evidence with the restored snapshot before forecasting.}
    \label{fig:figure3}
    \vspace{-10pt}
\end{figure*}

MissHyper targets the event initialization stage of a hypergraph IMTS forecaster. Before relational propagation begins, each sparse event node is enriched with two forms of layout-derived context: local observation support and co-timestamp evidence. We implement this snapshot-aware initialization with three encoder steps: mask-derived support encoding, timestamp-level synchronicity restoration, and missingness-guided fusion. Figure~\ref{fig:figure3} summarizes the pipeline.

This design focuses the proposed changes on representation initialization before downstream propagation. MissHyper uses only sparse-layout information, without imputing unmeasured values or converting sparse records into a dense observation grid.

Episodes can contain different numbers of observed events, and all event sets are processed in sparse form. The mask used for density estimation records only event availability and is not treated as a dense value input.

\subsection{Event-node formulation}

Each patient episode is represented as sparse timestamp-variable events. An observed event is \(e_n=(\tau_n,u_n,x_n)\), where \(\tau_n\), \(u_n\), and \(x_n\) denote timestamp, variable, and value. Forecast targets are represented as query events with known timestamp-variable positions but hidden values. This query-node view makes forecasting compatible with sparse relational learning: the model receives lookback observations and query locations, then estimates the hidden values attached to those query nodes.

Let \(\mathcal{O}_b\), \(\mathcal{Q}_b\), and \(\mathcal{V}_b=\mathcal{O}_b\cup\mathcal{Q}_b\) denote observed, query, and all event nodes for episode \(b\). Observed nodes carry measured values from the input window, while query nodes carry only timestamp and variable identity, with their values replaced by a neutral placeholder. A binary role flag prevents this placeholder from being confused with a true zero-valued observation. For \(n \in \mathcal{V}_b\),
\[
\tilde{x}_{b,n} =
\begin{cases}
x_{b,n}, & \text{if } n \text{ is an observed event},\\
0, & \text{if } n \text{ is a query event},
\end{cases}
\qquad
q_{b,n} =
\begin{cases}
0, & \text{if } n \text{ is an observed event},\\
1, & \text{if } n \text{ is a query event}.
\end{cases}
\]

This formulation keeps prediction local to nodes while preserving the distinction between evidence and targets. It requires only a timestamp index and a variable index for each node.

\subsection{Mask-derived support cue}

Clinical observations are unevenly distributed, so local event density provides a cue about how well an event is supported by nearby measurements. A value surrounded by many nearby records has stronger local support than the same value appearing after a long unobserved interval. MissHyper encodes this support as an explicit node-level cue before dependency learning.

Let \(\Omega_b \in \{0,1\}^{L \times C}\) be the structural event-position mask, with \(\Omega_{b,\ell,u}=1\) when position \((\ell,u)\) is an observed input position or a predefined forecast query position. Query coordinates are specified by the forecasting task, but their target values are never used. The mask is used only to compute a scalar support cue and is not passed to the forecasting backbone as a dense input sequence. For each position and node, we define
\[
\begin{aligned}
\rho_{b,\ell,u}
&=
\frac{1}{|\mathcal{W}_w(\ell)|}
\sum_{s\in\mathcal{W}_w(\ell)}
\Omega_{b,s,u},
\qquad
\rho_{b,\ell,u}\in[0,1],\\
\rho_{b,n}
&:=
\rho_{b,\tau_{b,n},u_{b,n}},
\end{aligned}
\]
where \(\mathcal{W}_w(\ell)\) is a clipped temporal window. Larger \(\rho\) indicates stronger nearby support, while \(1-\rho\) is the corresponding local missingness level. If a task does not expose query locations in advance, the same computation can be restricted to the observed prefix. The node input and initial embedding are
\[
\begin{aligned}
\mathbf{s}_{b,n}
&=
\left[
\tilde{x}_{b,n},
q_{b,n},
\rho_{b,n}
\right]^\top
\in \mathbb{R}^{3},\\
\mathbf{h}_{b,n}^{(0)}
&=
\phi_{\mathrm{in}}(\mathbf{s}_{b,n}),
\qquad
\mathbf{h}_{b,n}^{(0)}\in\mathbb{R}^{d},
\end{aligned}
\]
where \(\phi_{\mathrm{in}}\) is a learnable input projection.

This scalar cue helps distinguish observations with similar values but different local acquisition support, and later provides a reliability coordinate for gated fusion with same-timestamp context. We use a bounded average rather than a learned density estimator to keep this component simple and easy to isolate. The window size \(w\) controls the temporal scale of the cue and is selected on the validation split.

\subsection{Synchronicity restoration}

Flattening sparse records into event nodes can hide co-timestamp snapshots at initialization. In an ICU record, multiple variables measured at the same timestamp often correspond to a shared clinical recording event, such as vital-sign charting or a laboratory panel. Treating those measurements as isolated node inputs delays a directly observable relation until later message passing. MissHyper restores this structure before message passing by summarizing the observed evidence at each timestamp and broadcasting the summary to nodes with the same time index.

Let \(\pi_b(n)=\tau_{b,n}\) map a node to its timestamp and let \(\mathbf{H}^{(0)}_b\) stack initial node embeddings. We compute one support-weighted context per timestamp:
\[
\mathbf{c}_{b,\ell}
=
\frac{
\sum_{n\in\mathcal{O}_b:\tau_{b,n}=\ell} \rho_{b,n}\mathbf{h}^{(0)}_{b,n}
}{
\sum_{n\in\mathcal{O}_b:\tau_{b,n}=\ell} \rho_{b,n}+\epsilon
},
\qquad
\mathbf{C}_b=\operatorname{stack}_{\ell=1}^{L}\mathbf{c}_{b,\ell}.
\]
Here \(\epsilon\) handles empty timestamps. If timestamp \(\ell\) has no observed event, \(\mathbf{c}_{b,\ell}\) defaults to the zero vector before projection. Query-only timestamps therefore contain no target information and are handled by the same projection and gating mechanism. The contexts are projected and gathered back to event nodes:
\[
\widetilde{\mathbf{c}}_{b,n}
=
\phi_c\!\left(\mathbf{c}_{b,\pi_b(n)}\right),
\qquad
\widetilde{\mathbf{C}}_b
=
\operatorname{stack}_{n=1}^{|\mathcal{V}_b|}\widetilde{\mathbf{c}}_{b,n}.
\]
Thus each node receives co-timestamp context before downstream hypergraph propagation. We intentionally use deterministic support weighting rather than an additional attention layer: the goal is to expose a simple co-timestamp relation before downstream propagation, not to introduce another propagation-like module.

Only observed nodes contribute to the timestamp context. Query nodes may receive the restored context associated with their timestamp, but their hidden target values never enter the aggregation, preventing target-value leakage while still using known query locations. Support weighting gives larger influence to measurements with stronger local support, so the restored context combines instantaneous co-occurrence with local observation support before learned dependency propagation.

\subsection{Missingness-guided gated fusion}

Snapshot context should not be forced into every node equally. A sparse neighborhood may provide weak local evidence, in which case the gate can learn to borrow more from the restored timestamp context. A dense neighborhood provides stronger local support, allowing the update to remain closer to the node's own embedding when appropriate. Because the gate receives the node state, the timestamp context, and \(\rho\), the interpolation is explicitly conditioned on both representation content and local support. Uniformly replacing node features with a timestamp summary could wash out variable-specific evidence, while ignoring the summary would leave the initialization bottleneck unresolved.

MissHyper uses a dimension-wise gate conditioned on node state, timestamp context, and density. Let \(\boldsymbol{\rho}_b\in\mathbb{R}^{|\mathcal{V}_b|\times 1}\) stack node densities:
\[
\mathbf{A}_b
=
\sigma
\left(
\phi_g
\left(
\left[
\mathbf{H}^{(0)}_b
\,\Vert\,
\widetilde{\mathbf{C}}_b
\,\Vert\,
\boldsymbol{\rho}_b
\right]
\right)
\right),
\qquad
\mathbf{A}_b\in(0,1)^{|\mathcal{V}_b|\times d},
\]
where \(\Vert\) denotes concatenation and \(\sigma\) is sigmoid. The fused node representation is
\[
\mathbf{Z}^{(0)}_b
=
\mathbf{H}^{(0)}_b
+
\mathbf{A}_b\odot
\left(
\widetilde{\mathbf{C}}_b-\mathbf{H}^{(0)}_b
\right).
\]
The gate controls how far each node moves from local evidence toward restored timestamp context.

This residual interpolation keeps the fused representation in the same embedding space as the original node state. When a gate value is close to zero, the node remains dominated by its own input embedding. When a gate value is close to one, the node moves toward the restored timestamp context. Intermediate values allow the model to select different mixtures across hidden dimensions. In this way, missingness is not treated only as absence, but as a signal for how reliable isolated event evidence may be.

The fusion module is deliberately small: it creates no new edges and performs no iterative message passing. Its purpose is to prepare a better initial state for downstream dependency learning.

\subsection{Hypergraph forecasting backbone}

The context-aware embeddings \(\mathbf{Z}^{(0)}_b\) are passed to the unchanged hypergraph forecasting backbone. The backbone receives the same timestamp-node and variable-node incidence relations as the direct hypergraph baseline. Timestamp incidence connects records that share a time index, while variable incidence connects records from the same clinical variable across time. MissHyper changes neither relation. It only changes the node features entering the first propagation layer.
\[
\left(
\mathbf{Z}^{(K)}_b,
\mathbf{T}^{(K)}_b,
\mathbf{U}^{(K)}_b
\right)
=
\mathcal{B}_{\theta}
\left(
\mathbf{Z}^{(0)}_b,
\mathcal{A}^{t}_b,
\mathcal{A}^{u}_b
\right),
\]
where \(\mathcal{A}^{t}_b\) and \(\mathcal{A}^{u}_b\) are timestamp-node and variable-node incidence relations, and \(\mathbf{T}^{(K)}_b\), \(\mathbf{U}^{(K)}_b\) are final hyperedge states. Keeping the backbone architecture and propagation operators unchanged isolates the encoder-side contribution.

Conceptually, the backbone is responsible for higher-order dependency learning after initialization. The timestamp relation can refine synchronized measurements across layers, and the variable relation can exchange information along each clinical channel. MissHyper does not compete with these operations. Instead, it gives the first propagation layer node states that already contain a local snapshot estimate, reducing the need for early layers to spend capacity reconstructing same-timestamp evidence from scratch.

For query node \(n\), the decoder uses its final node state and incident hyperedge states:
\[
\mathbf{r}_{b,n}
=
\left[
\mathbf{z}^{(K)}_{b,n}
\,\Vert\,
\mathbf{t}^{(K)}_{b,\tau_{b,n}}
\,\Vert\,
\mathbf{u}^{(K)}_{b,u_{b,n}}
\right],
\qquad
\hat{x}_{b,n}=\psi_o(\mathbf{r}_{b,n}).
\]
With all mini-batch forecast queries collected in \(\mathcal{Q}\), training minimizes
\[
\mathcal{L}
=
\mathbb{E}_{(b,n)\in\mathcal{Q}}
\left[
\left\|
\hat{x}_{b,n}-x_{b,n}
\right\|_2^2
\right].
\]

The prediction head is applied only to query nodes. Observed nodes participate as evidence during propagation but do not contribute to the forecasting loss. Since MissHyper keeps the loss, decoder inputs, and backbone outputs unchanged, the training objective remains the standard multi-step forecasting objective used by the baseline. The additional computation is \(O(|V|w)\) for window-based density gathering and linear in the number of event nodes for timestamp aggregation and gated fusion. With the small validation-selected window sizes used in our experiments, this overhead remains limited relative to the hypergraph propagation backbone.

\section{Experiments}
\label{sec:experiments}

We evaluate whether snapshot-aware event initialization improves clinical IMTS forecasting under an unchanged hypergraph backbone architecture. The experiments examine three questions: whether MissHyper improves over representative irregular forecasting baselines, whether each encoder component contributes to the gain, and whether the support-density window is robust to reasonable choices.

\subsection{Experimental Setup}
\label{subsec:experimental_setup}

\paragraph{Datasets.}
We use three clinical IMTS benchmarks: PhysioNet 2012 (P12), MIMIC-III, and MIMIC-IV \citep{Silva2012PhysioNetChallenge,Johnson2016MIMICIII,Johnson2023MIMICIV}. PhysioNet 2012 provides ICU time-series records from the first 48 hours of patient stays and is processed at a 1-hour resolution. MIMIC-III is a large critical-care EHR benchmark whose ICU measurements are discretized into 30-minute intervals under the same setup. MIMIC-IV extends the MIMIC data ecosystem with more recent and denser ICU records, which are processed at a 1-minute resolution. Because clinical benchmark construction and extraction choices can affect time-series evaluation, we follow established preprocessing protocols for these datasets \citep{Purushotham2018BenchmarkingHealthcare,Harutyunyan2019MultitaskClinical,Wang2020MIMICExtract}. These datasets contain sparse and asynchronous ICU measurements with substantial variable-wise differences in recording frequency. As shown in Table~\ref{tab:dataset_stats_three}, the datasets differ in scale, variable dimensionality, observation density, and sequence length. This diversity allows us to test MissHyper across different sparsity and asynchrony regimes.

\begin{table}[t]
\centering
\scriptsize
\setlength{\tabcolsep}{2.5pt}
\renewcommand{\arraystretch}{0.98}
\caption{Dataset statistics for clinical IMTS forecasting. Avg. Obs. denotes the average number of observed values per patient episode.}
\label{tab:dataset_stats_three}
\vspace{2pt}
\begin{tabular*}{\linewidth}{@{\extracolsep{\fill}}lcccc}
\toprule
Dataset & Num. Vars & Num. Samples & Avg. Obs. & Max Length \\
\midrule
P12 & 36  & 11,981 & 308.6 & 47  \\
MIMIC-III          & 96  & 21,250 & 144.6 & 96  \\
MIMIC-IV           & 100 & 17,874 & 304.8 & 971 \\
\bottomrule
\end{tabular*}
\end{table}

\paragraph{Protocol.}
Each patient episode is represented as sparse timestamp-variable records rather than a dense time-by-variable grid. The model receives an observed prefix and predicts values at query positions within the forecasting horizon. Following prior clinical IMTS forecasting protocols \citep{Li2025HyperIMTS}, we split episodes into training, validation, and test sets with an 80\%/10\%/10\% ratio. Models are trained with shuffled mini-batches from the training split. The validation split is used only for early stopping and support-density window selection. For each seed, the checkpoint selected on the validation split is evaluated on the held-out test split. We use the dataset-specific discretization described above, following prior IMTS forecasting protocols.

\paragraph{Baselines and implementation.}
We compare MissHyper with representative methods for irregular multivariate time-series forecasting. Event- and attention-based baselines include PrimeNet, mTAN, and SeFT \citep{Chowdhury2023PrimeNet,Shukla2021MTAN,Horn2020SeFT}. Continuous-time and neural-flow baselines include NeuralFlows, CRU, and GNeuralFlow \citep{Bilos2021NeuralFlows,Schirmer2022CRU,Mercatali2024GNeuralFlow}. GRU-D represents missingness-aware recurrent modeling \citep{Che2018GRUD}. Graph-structured and recent irregular forecasting baselines include Raindrop, Warpformer, tPatchGNN, and GraFITi \citep{Zhang2022Raindrop,Zhang2023Warpformer,Zhang2024TPatchGNN,Yalavarthi2024GraFITi}. The most controlled comparison is the direct hypergraph baseline, where MissHyper uses the same construction and propagation operators but replaces the event initialization stage \citep{Li2025HyperIMTS}. All experiments are conducted on a Linux server with a single NVIDIA GeForce RTX 4090 GPU. The implementation uses PyTorch 2.9.1. We optimize all trainable parameters with Adam. The learning rate is kept at \(L_0\) for the first three epochs and then decayed as \(L_n = L_0 \times 0.8^{n-3}\) for epoch \(n>3\). Models are trained for up to 300 epochs with an early-stopping patience of 10 epochs. We report MSE and MAE as mean $\pm$ standard deviation over five random seeds, 2024--2028. For MissHyper, the batch size is 32 and the initial learning rate is \(5\times10^{-4}\). Backbone-related hyperparameters are kept fixed across the direct hypergraph comparison to isolate the effect of event initialization.

\subsection{Main Results}
\label{subsec:main_results}

\begin{table*}[t]
\centering
\scriptsize
\setlength{\tabcolsep}{2.5pt}
\renewcommand{\arraystretch}{0.98}
\caption{Multi-step forecasting performance (mean $\pm$ std over 5 runs). Lower is better.}
\label{tab:main_results}
\vspace{-1pt}
\begin{tabular*}{\textwidth}{@{\extracolsep{\fill}}lcccccc}
\toprule
 & \multicolumn{2}{c}{P12 (PhysioNet)} 
 & \multicolumn{2}{c}{MIMIC-III} 
 & \multicolumn{2}{c}{MIMIC-IV} \\
\cmidrule(lr){2-3} \cmidrule(lr){4-5} \cmidrule(lr){6-7}
Method & MSE & MAE & MSE & MAE & MSE & MAE \\
\midrule
PrimeNet     
& 0.7981 $\pm$ 0.0001 & 0.6872 $\pm$ 0.0002 
& 0.9105 $\pm$ 0.0002 & 0.6657 $\pm$ 0.0001
& 0.6682 $\pm$ 0.0001 & 0.5891 $\pm$ 0.0002 \\

NeuralFlows  
& 0.4098 $\pm$ 0.0031 & 0.4492 $\pm$ 0.0024 
& 0.6127 $\pm$ 0.0098 & 0.5342 $\pm$ 0.0061
& 0.4712 $\pm$ 0.0021 & 0.4782 $\pm$ 0.0012 \\

CRU          
& 0.6215 $\pm$ 0.0042 & 0.5791 $\pm$ 0.0035 
& 0.5921 $\pm$ 0.0101 & 0.5178 $\pm$ 0.0050
& 0.4368 $\pm$ 0.0019 & 0.4574 $\pm$ 0.0009 \\

mTAN         
& 0.3826 $\pm$ 0.0040 & 0.4318 $\pm$ 0.0031 
& 0.9473 $\pm$ 0.1092 & 0.6811 $\pm$ 0.0472
& 0.5361 $\pm$ 0.0118 & 0.5203 $\pm$ 0.0061 \\

SeFT         
& 0.7694 $\pm$ 0.0020 & 0.6728 $\pm$ 0.0030 
& 0.9286 $\pm$ 0.0017 & 0.6641 $\pm$ 0.0009
& 0.6735 $\pm$ 0.0003 & 0.5912 $\pm$ 0.0004 \\

GNeuralFlow  
& 0.8251 $\pm$ 0.0287 & 0.6812 $\pm$ 0.0094 
& 0.9012 $\pm$ 0.0186 & 0.6483 $\pm$ 0.0069
& 0.5023 $\pm$ 0.0017 & 0.4915 $\pm$ 0.0005 \\

GRU-D        
& 0.3467 $\pm$ 0.0025 & 0.4010 $\pm$ 0.0016 
& 0.4817 $\pm$ 0.0094 & 0.4563 $\pm$ 0.0053
& 0.6598 $\pm$ 0.0024 & 0.5834 $\pm$ 0.0015 \\

Raindrop     
& 0.3519 $\pm$ 0.0023 & 0.4087 $\pm$ 0.0021 
& 0.6782 $\pm$ 0.1714 & 0.5528 $\pm$ 0.0821
& 0.3429 $\pm$ 0.0038 & 0.3896 $\pm$ 0.0028 \\

Warpformer   
& 0.3089 $\pm$ 0.0014 & 0.3649 $\pm$ 0.0018 
& 0.4326 $\pm$ 0.0031 & 0.4047 $\pm$ 0.0016
& 0.2728 $\pm$ 0.0025 & 0.3129 $\pm$ 0.0020 \\

tPatchGNN    
& 0.3168 $\pm$ 0.0049 & 0.3715 $\pm$ 0.0046 
& 0.4465 $\pm$ 0.0107 & 0.4091 $\pm$ 0.0083
& 0.2761 $\pm$ 0.0019 & 0.3115 $\pm$ 0.0014 \\

GraFITi      
& 0.3098 $\pm$ 0.0017 & 0.3651 $\pm$ 0.0031 
& 0.4374 $\pm$ 0.0412 & 0.4148 $\pm$ 0.0285
& 0.2467 $\pm$ 0.0008 & 0.3018 $\pm$ 0.0005 \\

HyperIMTS 
& 0.3010 $\pm$ 0.0027 & 0.3619 $\pm$ 0.0045
& 0.4009 $\pm$ 0.0122 & 0.3745 $\pm$ 0.0099
& 0.2136 $\pm$ 0.0020 & 0.2793 $\pm$ 0.0020 \\

\midrule
MissHyper 
& \textbf{0.2962} $\pm$ 0.0008 & \textbf{0.3561} $\pm$ 0.0011
& \textbf{0.3860} $\pm$ 0.0033 & \textbf{0.3643} $\pm$ 0.0018
& \textbf{0.2081} $\pm$ 0.0025 & \textbf{0.2751} $\pm$ 0.0011 \\
\bottomrule
\vspace{-14pt}
\end{tabular*}
\end{table*}

Table~\ref{tab:main_results} reports the multi-step forecasting results on the three clinical benchmarks. MissHyper achieves the best MSE and MAE across all three datasets. Compared with the direct hypergraph baseline, MissHyper reduces MSE on all three datasets, from 0.3010 to 0.2962 on P12, 0.4009 to 0.3860 on MIMIC-III, and 0.2136 to 0.2081 on MIMIC-IV. The reductions are consistent across datasets, with the largest relative improvement on MIMIC-III. The parallel reductions in MAE indicate that the gains are not specific to MSE and reflect broader forecasting improvements.

The improvement is most pronounced on MIMIC-III, which has many variables but fewer average observations per episode, as shown in Table~\ref{tab:dataset_stats_three}. In this sparse high-dimensional setting, co-timestamp measurements are less frequent and easier to underrepresent during independent event initialization. When records are initialized independently, later layers must reconstruct more of the timestamp-level structure from sparse interactions. Restoring co-timestamp context before propagation therefore provides the downstream learner with a more informative starting point.

On MIMIC-III, MissHyper also reduces the MSE standard deviation from 0.0122 to 0.0033. This lower variance is consistent with the view that reducing the pre-propagation bottleneck can make downstream message passing less sensitive to random initialization.

\subsection{Ablation Study}
\label{subsec:ablation}

We ablate three encoder-side components: co-timestamp context restoration, adaptive fusion, and the support-density cue. All variants share the same propagation architecture and training protocol, so the comparison isolates the contribution of MissHyper's encoder design.

\begin{table*}[t]
\centering
\scriptsize
\setlength{\tabcolsep}{2.4pt}
\renewcommand{\arraystretch}{0.98}
\caption{Ablation study on multi-step forecasting (mean $\pm$ std over 5 runs). Lower is better.}
\label{tab:ablation}
\vspace{0pt}
\begin{tabular*}{\textwidth}{@{\extracolsep{\fill}}lcccccc}
\toprule
 & \multicolumn{2}{c}{P12 (PhysioNet)}
 & \multicolumn{2}{c}{MIMIC-III}
 & \multicolumn{2}{c}{MIMIC-IV} \\
\cmidrule(lr){2-3} \cmidrule(lr){4-5} \cmidrule(lr){6-7}
Variant & MSE & MAE & MSE & MAE & MSE & MAE \\
\midrule
MissHyper 
& \textbf{0.2962} $\pm$ 0.0008 & \textbf{0.3561} $\pm$ 0.0011
& \textbf{0.3860} $\pm$ 0.0033 & \textbf{0.3643} $\pm$ 0.0018
& \textbf{0.2081} $\pm$ 0.0025 & \textbf{0.2751} $\pm$ 0.0011 \\

w/o Snapshot Restoration 
& 0.2984 $\pm$ 0.0012 & 0.3587 $\pm$ 0.0015
& 0.3942 $\pm$ 0.0047 & 0.3710 $\pm$ 0.0026
& 0.2110 $\pm$ 0.0029 & 0.2783 $\pm$ 0.0016 \\

w/o Adaptive Gate 
& 0.2976 $\pm$ 0.0010 & 0.3579 $\pm$ 0.0013
& 0.3909 $\pm$ 0.0039 & 0.3676 $\pm$ 0.0022
& 0.2098 $\pm$ 0.0026 & 0.2766 $\pm$ 0.0014 \\

w/o Support-Density Cue 
& 0.2981 $\pm$ 0.0013 & 0.3584 $\pm$ 0.0016
& 0.3924 $\pm$ 0.0043 & 0.3691 $\pm$ 0.0025
& 0.2105 $\pm$ 0.0028 & 0.2774 $\pm$ 0.0015 \\
\bottomrule
\vspace{-26pt}
\end{tabular*}
\end{table*}

\begin{figure*}[t]
    \centering
    \begin{subfigure}[t]{0.49\textwidth}
        \centering
        \includegraphics[width=\linewidth]{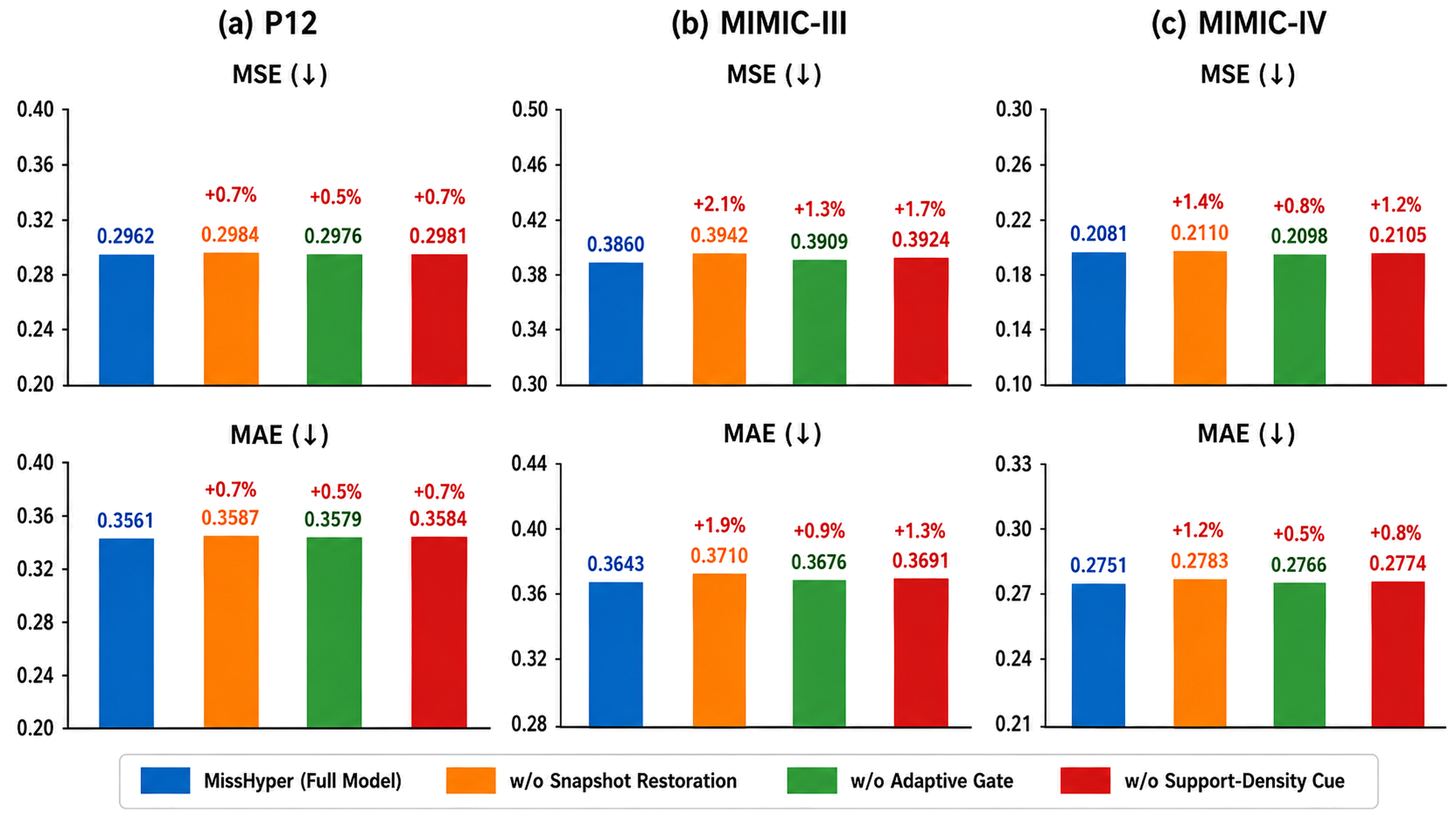}
        \caption{Ablation trends across datasets and metrics.}
        \label{fig:ablation_visual}
    \end{subfigure}
    \hfill
    \begin{subfigure}[t]{0.49\textwidth}
        \centering
        \includegraphics[width=\linewidth]{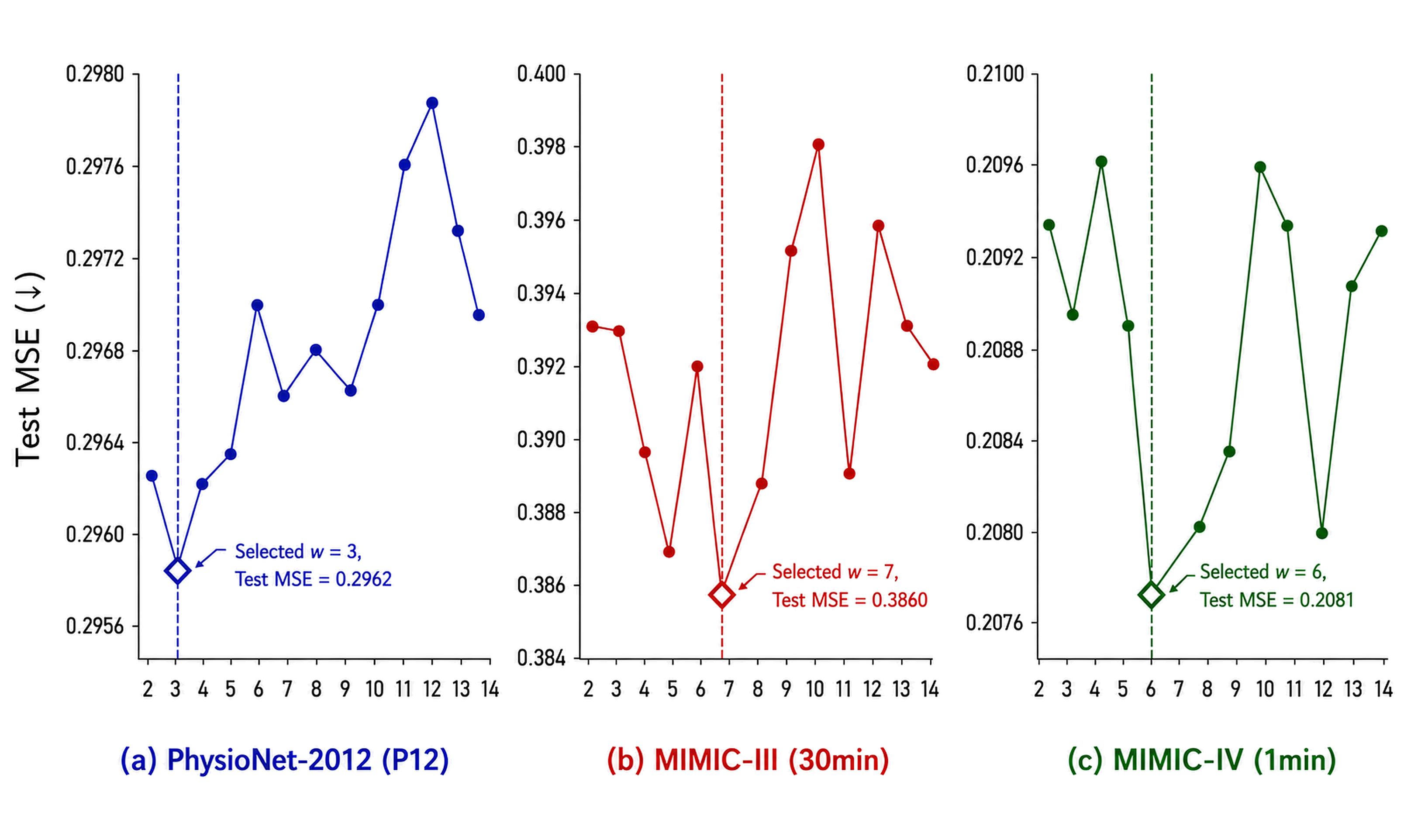}
        \caption{Support-density window sensitivity.}
        \label{fig:ws_sensitivity}
    \end{subfigure}
    \vspace{-6pt}
    \caption{Additional analyses for MissHyper. (a) Component-wise ablation verifies the contribution of snapshot restoration, adaptive fusion, and support-density encoding. (b) Window-size sensitivity shows robustness around the validation-selected support-density window.}
    \label{fig:additional_analysis}
    \vspace{-10pt}
\end{figure*}

The component-wise ablation trends are visualized in Figure~\ref{fig:ablation_visual}.

Table~\ref{tab:ablation} shows that MissHyper's gain comes from the combination of its encoder-side components rather than a single cue. All three components contribute to either restoring timestamp-level context or controlling how it is fused into event representations. The largest degradation occurs without co-timestamp context restoration, supporting the central role of pre-propagation synchronicity. On MIMIC-III, removing this component increases MSE from 0.3860 to 0.3942. Removing the adaptive gate also degrades performance, indicating that restored context should be fused selectively rather than applied uniformly. Finally, removing the support-density cue consistently hurts performance, suggesting that local missingness provides reliability information beyond the raw value and role indicator.

\subsection{Sensitivity to the Support-Density Window}
\label{subsec:window_sensitivity}

We select the support-density window size using only the validation split by sweeping \(w\in\{2,3,\ldots,14\}\). The selected values are \(w=3\) for P12, \(w=7\) for MIMIC-III, and \(w=6\) for MIMIC-IV. These values are fixed before test evaluation and used for the main results in Table~\ref{tab:main_results}.

Figure~\ref{fig:ws_sensitivity} reports test performance for nearby window choices after the validation-selected values are fixed. These additional test points are reported only for robustness inspection and are not used to select \(w\). Nearby choices yield similar test performance, indicating that the method is reasonably robust to this hyperparameter.

The selected values also reflect dataset-specific time units: P12 uses 1-hour bins, so \(w=3\) covers about 3 hours, while MIMIC-III uses 30-minute bins and \(w=7\) covers about 3.5 hours. MIMIC-IV is discretized at 1 minute, so \(w=6\) corresponds to a shorter physical interval, consistent with the finer temporal granularity of this processed benchmark.

\section{Conclusion and Discussion}

We studied a pre-propagation representation bottleneck in event-centric clinical IMTS models: co-timestamp records are often initialized independently even when they jointly form a patient-state snapshot. MissHyper addresses this issue by augmenting sparse event nodes with a local support-density cue, restoring timestamp-level context, and using a missingness-guided gate to fuse this context with node-specific evidence. Experiments on P12, MIMIC-III, and MIMIC-IV show consistent gains, with clear improvements over the direct hypergraph baseline, supporting the view that event representations should reflect not only measured values but also the acquisition context in which those values appear.

More broadly, our results suggest that sparse clinical forecasting benefits from treating missingness and co-timestamp structure as part of the representation problem, rather than as signals left to later dependency propagation. This perspective may also inform other event-centric models in which observations are sparse, asynchronous, and shaped by domain-specific measurement processes.

MissHyper has several limitations. It uses a fixed local window to estimate support density and focuses on numerical clinical time series without incorporating additional EHR modalities. It also relies on hypergraph attention operations, whose scalability may become a concern for very long sequences or high-dimensional records. Although methods such as MissHyper could support earlier risk monitoring from sparse ICU records, inaccurate forecasts, dataset bias, privacy constraints, and over-reliance on automated predictions remain important risks in deployment. Any clinical use would require external validation, calibration checks, privacy-preserving data handling, and careful subgroup monitoring.

Future work may explore adaptive density estimators, multi-scale synchronicity restoration, and extensions beyond hypergraph backbones, including joint forecasting-imputation and downstream clinical prediction tasks.

\bibliographystyle{unsrtnat}
\bibliography{misshyper_references}


\end{document}